# 6D Pose Estimation using an Improved Method based on Point Pair Features

Joel Vidal, Chyi-Yeu Lin
Department of Mechanical Engineering
National Taiwan University of Science and Technology
Taipei, Taiwan
e-mail: jolvid@gmail.com; jerrylin@mail.ntust.edu.tw

Robert Martí
Computer Vision and Robotics Group
University of Girona
Girona, Spain
e-mail: robert.marti@udg.edu

*Abstract*—The Point Pair Feature [4] has been one of the most successful 6D pose estimation method among model-based approaches as an efficient, integrated and compromise alternative to the traditional local and global pipelines. During the last years, several variations of the algorithm have been proposed. Among these extensions, the solution introduced by Hinterstoisser et al. [6] is a major contribution. This work presents a variation of this PPF method applied to the SIXD Challenge datasets presented at the 3rd International Workshop on Recovering 6D Object Pose held at the ICCV 2017. We report an average recall of 0.77 for all datasets and overall recall of 0.82, 0.67, 0.85, 0.37, 0.97 and 0.96 for hinterstoisser, tless, tudlight, rutgers, tejani and doumanoglou datasets, respectively.

## I. Introduction

3D Object Recognition and specially 6D pose estimation problems are crucial steps in object manipulation tasks. During the last decades, 3D data and feature-based methods have gained reputation with a wide range of presented model-based approaches [13]-[17].

In general, the model-based methods are divided in two main pipelines: Global and Local. Global approaches [14]-[17] describe the whole object or its parts using one global description. Local approaches [13] describe the object by using local descriptions around specific points. Global descriptions usually require the segmentation of the target object or its parts and they tend to ignore local details which may be discriminative. These characteristics make global approaches not robust against occlusion and highly cluttered scenes. On the other hand, local approaches are usually more sensitive to sensor noise due to their local nature and they tend to show a lower performance on symmetric object or objects with repetitive features.

On this direction, the Point Pair Features approach, proposed by Drost et al. [4], has been proved [9] one of most successful methods showing strong attributes as a compromise solution that integrates benefits of both local and global approaches. Among several proposed extensions of the method, Hinterstoisser et al. [6] analyzed some of the weakest points and proposed an extended solution that provided a significant improvement in presence of sensor noise and background clutter. This paper proposes a new variation of this method and test its performance against the challenging set of datasets presented recently at the SIXD Challenge 2017 [1] organized at the 3rd International Workshop on Recovering 6D Object Pose at ICCV 2017.

## II. The Point Pair Features Approach

The proposed method follows the basic structure of the Point Pair Feature(PPF) approach defined by Drost et al. [4] which consists of two stages: global modeling and local matching. The main idea behind this approach is to find, for each scene point, the corresponding model point and their rotation angle, defining a transformation from two points, their normal and the rotation around the normal. This correspondence is found by using a four-dimensional feature (fig. 1), defined between every pair of two points and their normals, so that each model point is defined by all the pairs created by itself and all the other model points.

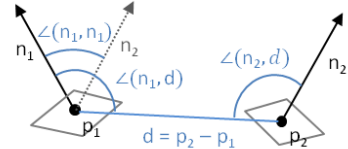

$$F(p_1, p_2, n_1, n_2) = (\ \|d\|_2, \angle(n_1,d), \angle(n_2,d), \angle(n_1,n_2)\ )$$

Figure 1. Point Pair Feature definition

Initially, on the global modeling stage, the input model data is preprocessed by subsampling the data. Then, a 4-dimensional lookup table storing the model pairs is constructed by using the discretized PPF as an index (fig. 2). This table will provide a constant access to all the model correspondence reference points and their rotation angles for each cell pointed by a discretized PPF feature obtained from a scene pair.

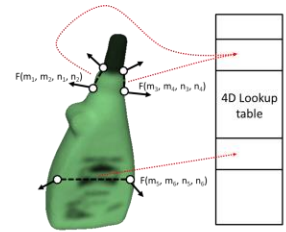

Figure 2. Example of global modeling. Three pairs of points are stored in the 4D Lookup table using their discretized PPF feature as an index.

During the local matching stage, the input data is preprocessed using the same techniques as the modeling part. For each given scene point, all the possible PPF are discretized and used as an index of the lookup table, obtaining a set of pairs of model points and rotation angles representing all the possible corresponding candidates. Each of these candidates casts a vote on a table, in a Hough-like voting scheme, where each value represents a hypothesis transformation defined by a model point and a rotation angle (fig. 3). Then, the peak value will be extracted as the best candidate for this scene point correspondence. Finally, all the hypotheses obtained from the scene points are clustered and a set of postprocessing steps are applied to extract the best one.

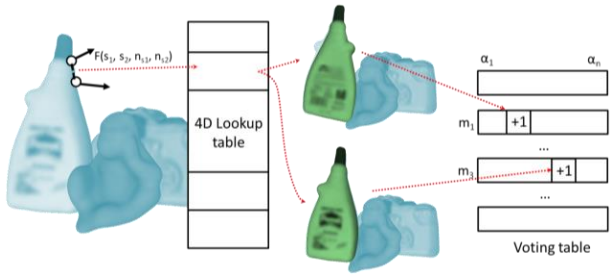

Figure 3. Example of local matching. For a point of the scene, a given pair is matched using the 4D lookup table obtaining two corresponding candidates, which defines two hypothesis transformations. Each candidate cast a vote on the voting table.

### III. IMPROVEMENTS

Following the analysis and reasoning introduced by [6], we propose a set of new improvements for the basic PPF method.

On the preprocessing part, during the subsampling step, the point pairs with angles between normals larger than 30 degrees are kept, but using a local clustering approach instead. In addition, after the clustering, an additional filtering step is proposed to remove non-discriminative pairs between neighboring clusters. In practice, this clustering area size is the same as the subsampling step.

To improve the run-time performance of the method during the local matching stage, only the points pairs with a distance smaller than the model diameter are checked by using a kd-tree structure. Following the ideas proposed by [6] the system avoids casting a vote twice for the same discretized PPF and rotation angle and also checks all the PPF index neighbors to account for sensor noise. Instead of checking all 80 neighbors in the lookup table, we propose a more efficient solution, by only voting for those neighbors that have big chances to be affected by noise checking the quantization error (fig. 4).

After clustering the hypothesis, a simplified view-dependent re-scoring process is used on the most voted 500 hypothesis. In this process, the hypotheses are reordered based on how well they fit to the scene data. In addition, to improve the score robustness, a Projective ICP [11] refinement is performed on the first 200.

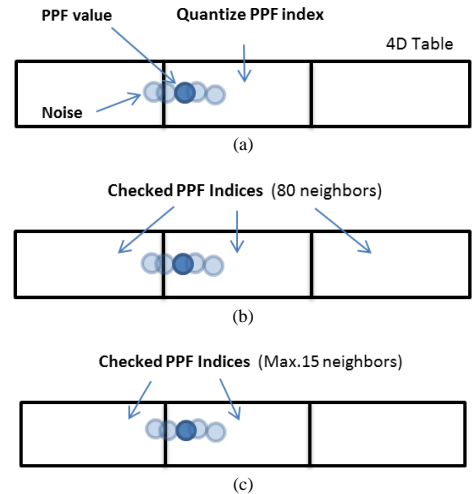

Figure 4. Scheme of the neigbhour cheking to account for sensor noise during quantization. (a) On dimension representation of the noise effect during quantization of a PPF value. (b) Solution proposed by Hinterstoisser at al. [6] checking the 80 niegbors of the 4D table. (c) Our proposed solution, checking up to 15 neigbors of the 4D table.

Finally, two filtering postprocessing steps are applied to discard special ambiguous cases such as planes and partially matching surfaces. The first step checks for non-consistent points removing hypothesis that are partially fitting the scene but do not have enough consistence with the scene points. A second step checks the overlapping rate of the object silhouette with respect to the scene edges in order to filter well-fitting objects with non-matching boundaries.

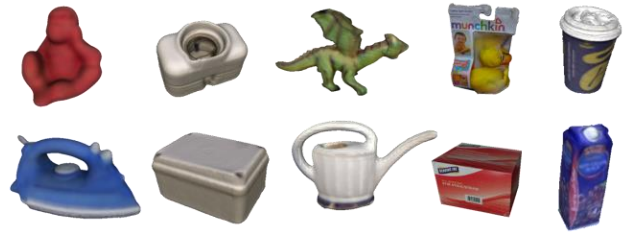

Figure 5. Some of the models used in the datasets

TABLE I. DATASETS MODELS AND RGB-D TEST IMAGES

| Datasets | Models | Test images |
| --- | --- | --- |
| Hinterstoisser et al. [5][2] | 15 | 18273 |
| T-Less [7] | 30 | 10080 |
| TUD Light | 3 | 23914 |
| Rutgers APC [10] | 14 | 5964 |
| Tejani et al. [12] | 6 | 2067 |
| Doumanoglou et al. [3] | 2 | 177 |

## IV. DATASETS

The SIXD Challenge 2017 [1] proposed a set of datasets for evaluating the task of 6D localization of a single instance of a single object. The mentioned datasets, shown in Table 1, are *hinterstoisser*, *tless*, *tudlight*, *rutgers*, *tejani* and *doumanoglou*. Each dataset contains a set of 3D objects models and RGB-D test images. The proposed scenes cover a wide range of cases with a variety of objects in different poses and environments including multiple instances, clutter and occlusion. The six datasets contain a total of 68 different object models (fig. 5) and 60475 test images. Notice that *rutgers*, *tejani* and *doumanoglou* are reduced versions and the *doumanoglou*'s models are also included in *tejani*.

## V. RESULTS

The proposed method was evaluated using the Visible Surface Discrepancy (VSD) [8] metric setting the parameters δ, τ and $t$ to 15 mm, 20 mm and 0.35, respectively. All datasets were evaluated using the same parameters for all models and scenes.

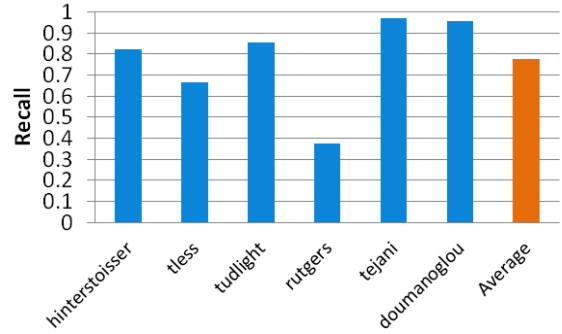

Figure 6. Recall rate for each dataset and total average

Specifically, the used subsampling leaf was defined as 0.05 of the model diameter and the PPF quantization was set to 20 divisions for distance and 15 divisions for angles.

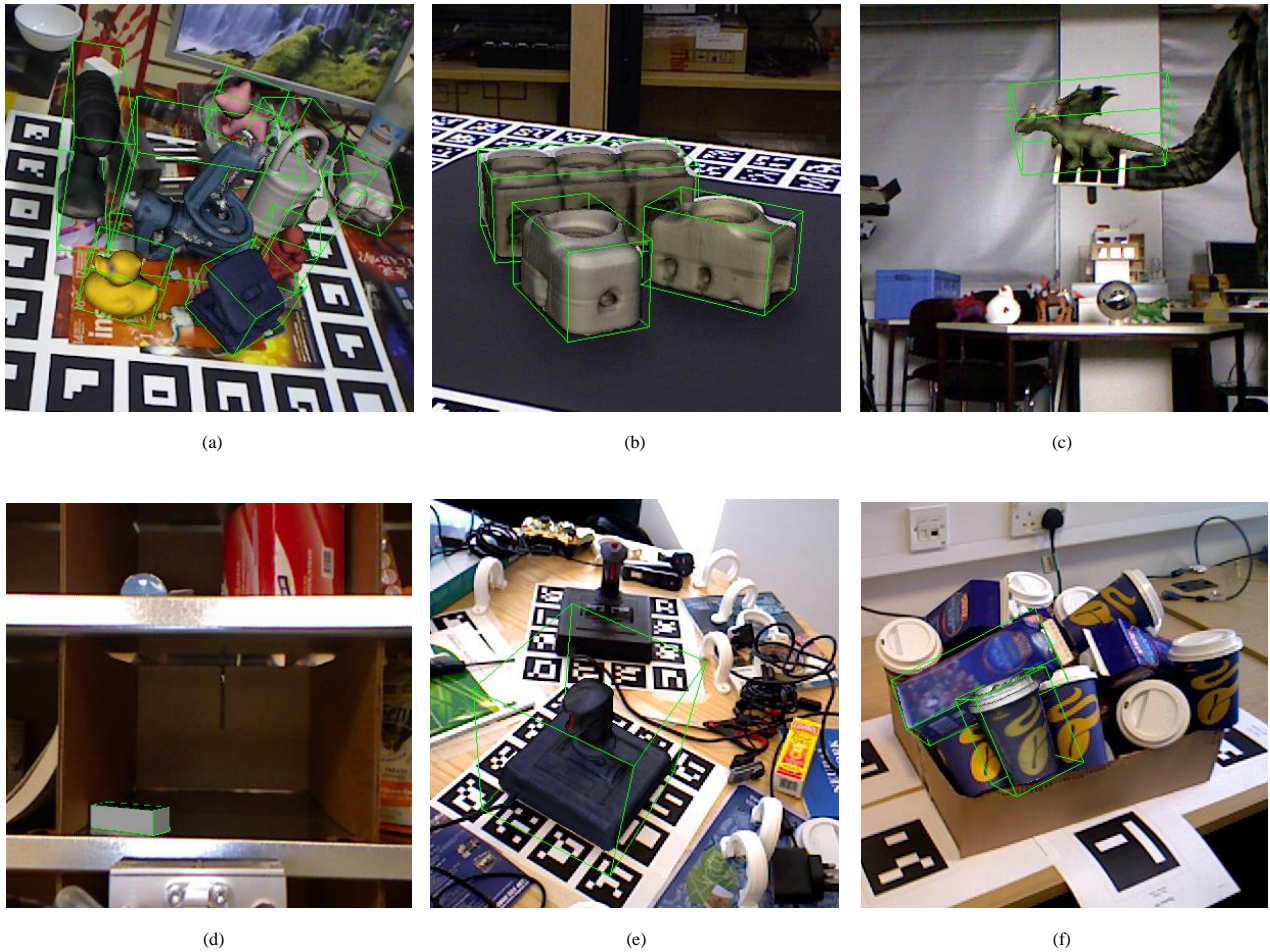

Figure 7. Exemples of results for all datasets: (a) hinterstoisser, (b) tless, (c) tudlight, (d) rutgers, (e) tejani and (f) doumanoglou.

The obtained overall recall rate for each dataset and the total average recall are shown in fig. 6. In addition, the individual recall rate for each of the models for all datasets is shown in fig. 8.

The results show a high and consistent performance across most datasets with an average recall rate of 0.77. The *tejani* and *doumanoglou* datasets shows the highest overall recognition rate with a recall value of 0.97 and 0.96 respectively. *Tudlight, hinterstoisser* and *tless* shows a recall rate of 0.85, 0.82 and 0.67. Finally, with the lowest recognition performance, the *rutgers* dataset shows an overall recall rate of 0.37. *Rutgers* is a challenging dataset as their CAD object models with few features are easily mismatched with the bookshelf-like structure scenes (fig. 7d), making more challenging the extraction of contour information and increasing the false negative cases.

Figure 7 shows an example of the results obtained for all datasets.

## VI. CONCLUSION

This work proposes a new improved variation of the PPF approach and tests its performance against the recently released set of datasets introduced at the SIXD Challenge 2017 [1] including 68 object models and 60475 test images.

The proposed method introduces a novel subsampling step with normal clustering and neighbor pairs filtering, in addition to a faster kd-tree neighbor search and a more efficient solution to decrees the effect of the sensor noise during the quantization step. Finally, the method uses several postprocesing steps for re-scoring, refining and filtering the final hypothesis.

The obtained results, using the VSD [8] metric, show a high and consistent performance across most datasets with an average recall of 0.77, with the exception of *rutgers* dataset, which shows a significant lower rate.

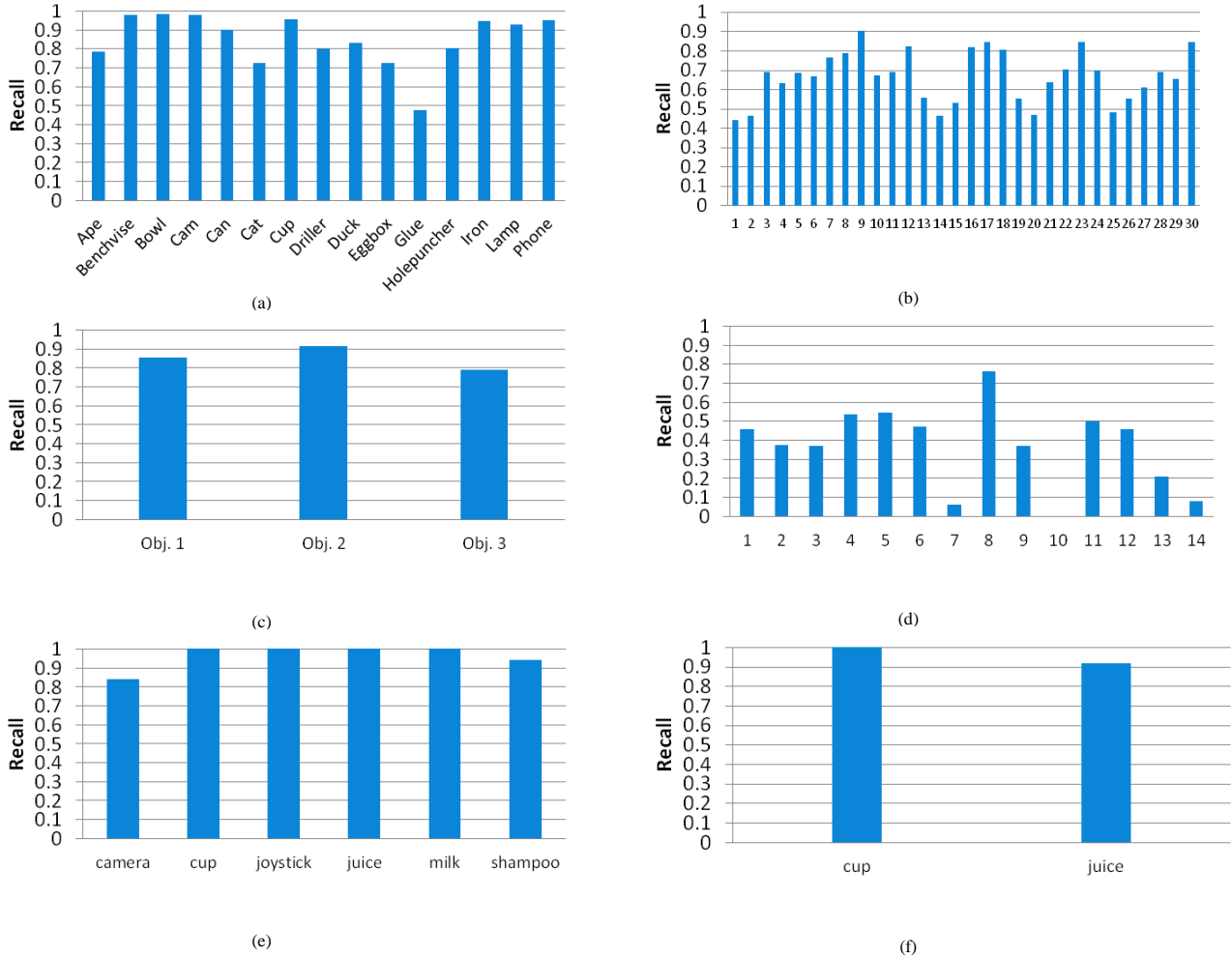

Figure 8. Recall rate for each model of all the datasets: (a) hinterstoisser, (b) tless, (c) tudlight, (d) rutgers, (e) tejani and (f) doumanoglou.

## REFERENCES

[1] SIXD Challenge 2017. http://cmp.felk.cvut.cz/ sixd/challenge_2017/. Accessed: 2017-9-28.

[2] E. Brachmann, A. Krull, F. Michel, S. Gumhold, J. Shotton, and C. Rother, "Learning 6D Object Pose Estimation Using 3D Object Coordinates," In Proceedings of the European Conference on Computer Vision (ECCV), 2014.

[3] A. Doumanoglou, R. Kouskouridas, S. Malassiotis, and T. K. Kim, "Recovering 6d object pose and predicting next-best view in the crowd," In The IEEE Conference on Computer Vision and Pattern Recognition (CVPR), June 2016.

[4] B. Drost, M. Ulrich, N. Navab, and S. Ilic, "Model globally, match locally: Efficient and robust 3d object recognition," In 2010 IEEE Conference on Computer Vision and Pattern Recognition (CVPR), June 2010, pp. 998–1005.

[5] S. Hinterstoisser, V. Lepetit, S. Ilic, S. Holzer, G. Bradski, K. Konolige, and N. Navab, "Model Based Training, Detection and Pose Estimation of Texture-Less 3D Objects in Heavily Cluttered Scenes," Springer Berlin Heidelberg, Berlin, Heidelberg, 2013, pp. 548–562.

[6] S. Hinterstoisser, V. Lepetit, N. Rajkumar, and K. Konolige, "Going Further with Point Pair Features," In Proceedings of the European Conference on Computer Vision (ECCV), 2016.

[7] T. Hodan, P. Haluza, . Obdrlek, J. Matas, M. Lourakis, and X. Zabulis, "T-LESS: An RGB-D Dataset for 6d Pose Estimation of Texture-Less Objects," In 2017 IEEE Winter Conference on Applications of Computer Vision (WACV), Mar. 2017, pp. 880–888.

[8] T. Hodan, J. Matas, and S. Obdrzalek, "On Evaluation of 6D Object Pose Estimation," In ECCV Workshop, 2016

[9] L. Kiforenko, B. Drost, F. Tombari, N. Krger, and A. Glent Buch, "A performance evaluation of point pair features," Computer Vision and Image Understanding, Sept. 2017.

[10] C. Rennie, R. Shome, K. E. Bekris, and A. F. D. Souza, "A dataset for improved rgbd-based object detection and pose estimation for warehouse pick-and-place," IEEE Robotics and Automation Letters, July 2016, 1(2):1179–1185.

[11] S. Rusinkiewicz and M. Levoy, "Efficient variants of the ICP algorithm," In Proceedings Third International Conference on 3-D Digital Imaging and Modeling, 2001, pp. 145–152.

[12] A. Tejani, D. Tang, R. Kouskouridas and T.-K. Kim, "Latent-class hough forests for 3d object detection and pose estimation," In European Conference on Computer Vision, 2014, pp. 462–477.

[13] Y.Guo, M. Bennamoun, F. Sobel, M. Lu, J. Wan, "3D Object Recognition in Cluttered Scenes with Local Surface Features: A Survey," IEEE Transactions on Pattern Analysis and Machine Intelligence, 2014, 36(11) :2270-2287.

[14] R. Rusu, G. Bradski, R. Thibaux, and J. Hsu, "Fast 3D recognition and pose using the viewpoint feature histogram," in Proc. IEEE/RSJ Int. Conf. Intell. Robots Syst., 2010, pp. 2155–2162.

[15] R. Osada, T. Funkhouser, B. Chazelle, and D. Dobkin, "Shape distributions," ACM Trans. Graphics, vol. 21, no. 4, pp. 807–832, 2002.

[16] E. Wahl, G. Hillenbrand, and G. Hirzinger. Surflet-pair relation histograms: A statistical 3d-shape representation for rapid classification. In 3-D Digital Imaging and Modeling, 3DIM 2003. Proceedings. Fourth International Conference on, pp. 474–481, 2003

[17] R. Schnabel, R. Wahl, and R. Klein. Efficient RANSAC for point-cloud shape detection. In Computer Graphics Forum, vol. 26, pp. 214–226. 2007.